\documentclass[letterpaper, 10 pt, conference]{ieeeconf}  %

\IEEEoverridecommandlockouts                              %

\usepackage{amsmath}

\usepackage{accents}

\makeatletter
\def\BState{\State\hskip-\ALG@thistlm}
\makeatother

\usepackage{amssymb}

\usepackage{algorithm}
\usepackage[noend]{algpseudocode}

\usepackage{mathrsfs}
\usepackage{mathtools}
\usepackage{amsfonts}

\usepackage{subcaption}

\usepackage{stackengine}
\setstackgap{S}{2pt}

\usepackage[
    colorlinks,
    urlcolor=black,
    citecolor=black,
    linkcolor=black,
    pdfpagemode=UseOutlines, %
  ]{hyperref}
  
\usepackage[capitalise]{cleveref}
\usepackage{xcolor}

\usepackage{siunitx}

\usepackage{multirow}
\usepackage{cite}
\usepackage{balance}
\usepackage{url}

\newcommand{\RRR}{{\mathbb{R}}}
\newcommand{\XX}{{\mathcal{X}}}
\newcommand{\1}{{\text{-1}}}

\title{\LARGE \bf
Regrasp Maps for Sequential Manipulation Planning 
}

\author{Svetlana Levit$^{1, 2}$ and Marc Toussaint$^{1, 2}$%
\thanks{This research has been supported by the German Research Foundation (DFG) under Germany's Excellence Strategy – EXC2002/1–390523135 "Science of Intelligence"}%
\thanks{$^{1}$Learning \& Intelligent Systems Lab, TU Berlin, Germany
        }%
\thanks{$^{2}$Science Of Intelligence Cluster of Excellence}%
}

\begin{document}

\maketitle
\thispagestyle{empty}
\pagestyle{empty}

\begin{abstract}

We consider manipulation problems in constrained and cluttered settings, which require several regrasps at unknown locations. We propose to inform an optimization-based task and motion planning (TAMP) solver with possible regrasp areas and grasp sequences to speed up the search.
Our main idea is to use a state space abstraction, a \emph{regrasp map}, capturing the combinations of available grasps in different parts of the configuration space, and allowing us to provide the solver with guesses for the mode switches and additional constraints for the object placements.
By interleaving the creation of regrasp maps, their adaptation based on failed refinements, and solving TAMP (sub)problems, we are able to provide a robust search method for challenging regrasp manipulation problems.
\end{abstract}

\section{Motivation}
In a scene with obstacles, objects and free space, Task-and-Motion Planning (TAMP) usually considers the movable obstacles and objects during the task planning stage and accounts for geometric constraints of collision-free motion during the trajectory search.
However, \emph{temporal inter-dependencies} of geometrical and symbolical constraints can prevent the planner from finding a feasible multi-step solution.
For instance, a common issue arises
when the planner suggests a placement location, which does not allow an object to be picked in a collision-free manner by the grasp needed for the next step.
An established method to partition the space to account for obstacles in motion planning is using the occupancy maps.
Motivated by the idea of creating a map for sequential manipulation planners, we present an abstraction to partition the free object configuration space into areas of similar grasp signatures, allowing us to plan for transitions that account for multi-step interdependent geometric constraints.

\begin{figure}[t]   
  \subfloat[Cage problem]{\label{fig:gridsamples}\includegraphics[width=0.49\linewidth]{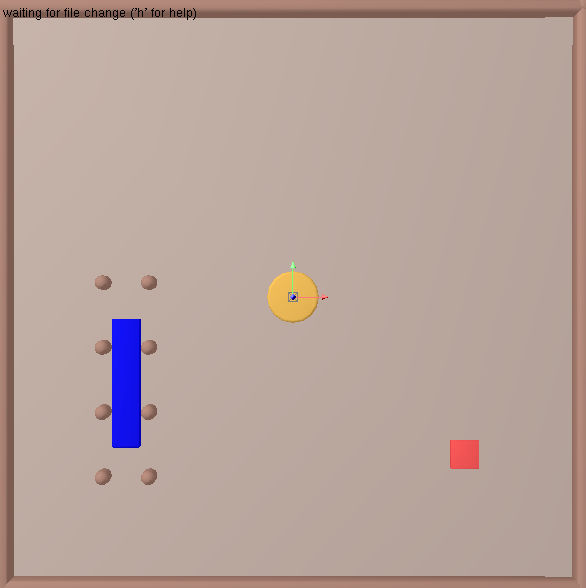}}\hfill
  \subfloat[Segmented scene and a sequence of abstract regrasp states]{\label{fig:path}\includegraphics[width=0.49\linewidth]{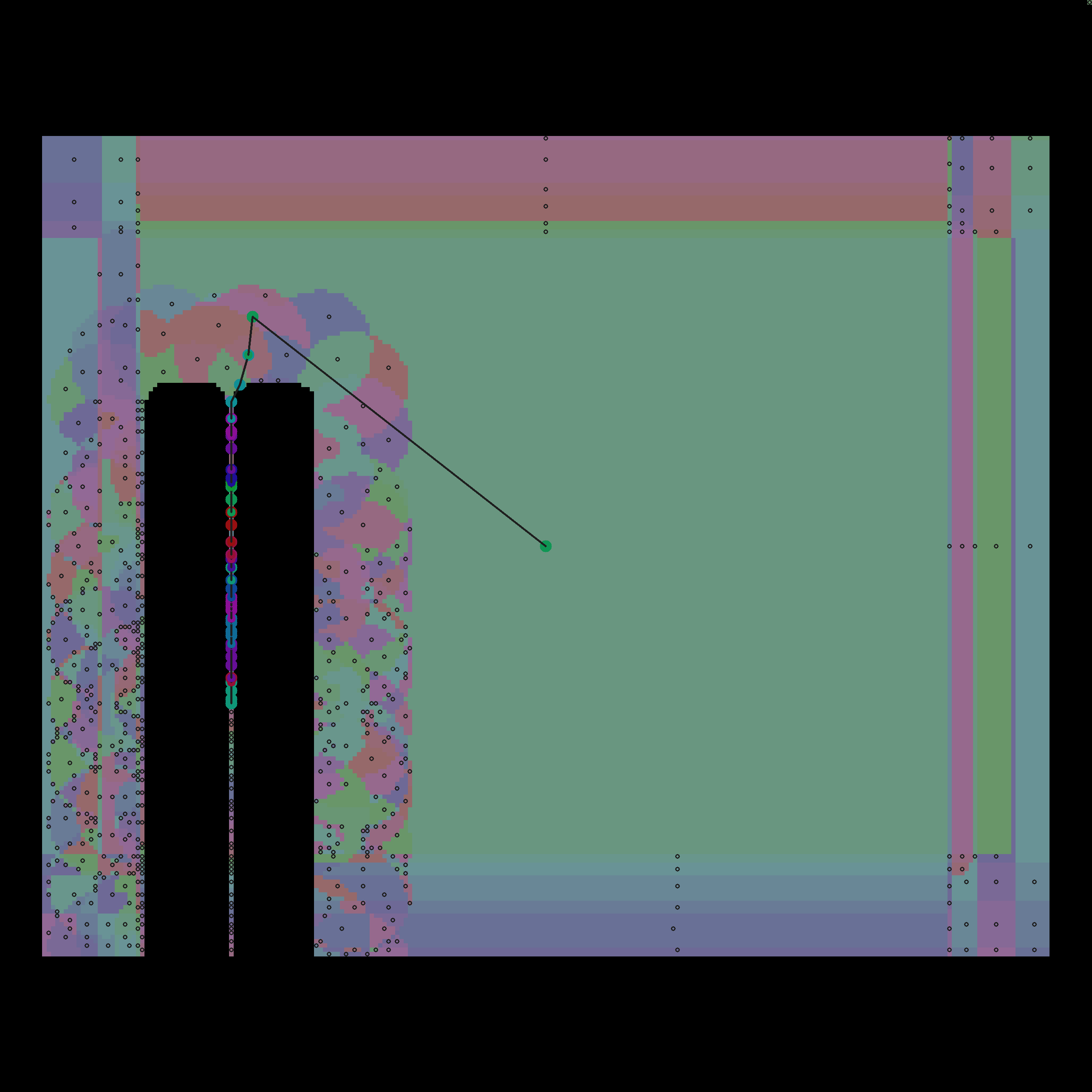}}\hfill
\caption{An abstract regrasp plan: the agent (yellow circle) needs to move the blue stick to the red goal. A sequence of regrasps (\ref{fig:path}) is needed to get the stick out of the cage first. The segmentation indicates which object placements would allow the same grasps, with parts not collision-free for the object shown in black.}
\label{fig:sticksolution}
\end{figure}

In our work we address the following challenges of sequential manipulation with regrasps:
\begin{itemize}
\item \emph{Strong interdependency} between grasps and placements over several steps.
\item Requirement for \emph{precise grasps and placements}, which are difficult to sample.
\item \emph{Narrow set of feasible grasps} in certain steps, making multi-step solutions hard to discover.
\end{itemize}
Figure~\ref{fig:sticksolution} shows an example problem, inspired by~\cite{Simeon2004}, where a long blue stick is trapped between the bars and to move it to the goal location (red square), the agent (in yellow) needs to move the stick in a sequence of short moves and regrasps.
At every position of the object the surrounding obstacles limit possible grasps. Our approach relies on segmenting the object configuration space based on available grasps, the \emph{grasp signature}, and thus providing our planner with a map of possible regrasps. Fig.\,\ref{fig:path} visualizes the segmentation results and a possible path in the abstract regrasp map with the grasp sequence shown as circles of different colors.
Our approach uses the information from the regrasp map to find feasible object transport trajectories by refining the abstract plans to geometrically valid solutions. To integrate the regrasp information, we use $k$-Order Markov
Optimization (KOMO)~\cite{toussaint2014newtonmethodskordermarkov, Toussaint2015 } constraint formulation and represent solutions as action skeletons. %
By informing the solver about the grasps sequences, grasp feasibility measures, and intermediate placements, we reduce the number of nonlinear
mathematical programs (NLPs) needed to be solved and address the regrasp challenges. %
In summary, our main contributions are:
\begin{itemize}
\item We propose a state space abstraction, a \emph{regrasp map}, providing an optimization-based TAMP solver with additional constraints for solving sequential manipulation problems with regrasps.
\item We develop an adaptive refinement method interleaving regrasp map search and solving of TAMP (sub)problems to account for failed refinements.
\item We empirically demonstrate the effectiveness of these abstract regrasp sequences and grasp-feasibility information on a set of long-horizon regrasp problems.
\end{itemize}

\section{Related Works}
\subsection{Using Abstract Graphs for Planning}
Many works incorporate graphs into motion or manipulation planning to guide solvers based on previous solutions. %
Kingston \emph{et al.}~\cite{Kingston_2021} construct roadmaps in an augmented foliated space for multi-modal manipulation. Their approach is learning inside one mode and using experience to bias a sampling-based algorithm for problems in the same mode family. 

Other approaches use graphs derived from \emph{skills} or \emph{previous solutions}. For example, Bagaria \emph{et al.}~\cite{Bagaria21a} build graphs whose nodes are skill subgoals and edges are skill policies learned from similar problems. Closer to our setting, Elimelech \emph{et al.}~\cite{Elimelech_2023} propose \emph{abstract state road maps} (ARM), derived from previous task solutions, to guide new tasks. In contrast, we do not rely on previously learned skills or solutions, making our approach directly applicable to new problems.

ALEF framework~\cite{Kingston_2023a} reuses experience from similar modes in the form of discrete leads biasing the search toward useful mode transitions. %
Similarly to our work, ALEF models the success probability of transitions, but the graph states are derived from the problem symbols, while we rely on geometry of the scene and available grasps.

\subsection{Optimization-based TAMP}
The idea to use weighted connections between modes for an optimization-based planner was explored in~ \cite{Toussaint2018, Phiquepal2019}, which combines an LGP (Logic Geometric Program) solver and an RRT (Rapidly-Exploring Random Trees) motion planner for manipulation with a discrete set of modes. In~\cite{Semenov2024} the authors use a reachability graph as a heuristic for an LGP planner during path planning as a guidance to overcome local minima caused by obstacles; however, their graph does not consider manipulation constraints.

\subsection{Multi-Modal Planning}
Mode-transition graphs are widely used in multi-modal planning \cite{Bretl_2006, Mirabel_2016, Mirabel_2017, Schmitt_2019, Asselmeier24}, including earlier formulations with PRMs\cite{Nielsen_2000}, and partitioning the configuration space based on possible (continuous and discrete) mode families in \cite{Hauser2011}.
Simeon \emph{et al.}~\cite{Simeon2004} proposed a \emph{manipulation roadmap} to plan multi-step transit and transfer actions, finding connected components by sampling the object’s configuration space. Our approach similarly partitions the configuration space, but specifically captures \emph{grasp signatures} that enable or restrict regrasps in narrow areas to account for multi-step inter-dependencies.

\subsection{Finding Grasps and Grasp Sequences}
Complementary to our approach is the topic of grasp synthesis and optimal grasp selection, where researchers commonly make use of 3D data to predict reachable, collision-free grasps in clutter~\cite{Hoang22, Sundermeyer21, Xibai2020}. For sequential manipulation tasks, a related challenge is choosing and chaining multiple actions. Hu \emph{et al.}~\cite{Hu_2023}  models the manipulations as manifolds for sliding, grasping, and transfer to handle grasping in difficult settings. For tasks requiring object reorientation, ~\cite{wan2018auro, Ma2019RegraspPU} search for regrasp sequences taking into account accessible grasps at stable intermediate placements.   %

\section{Problem Formulation and Definitions}
\label{sec:formulation}
\subsection{Logic Geometric Programming for TAMP}
We consider the task of moving a single target object to a goal area. The problem's configuration space $\XX=\RRR^n \times SE(3)$ is defined by the $n$-DoF robot and the position of the movable object. 

We want to find a trajectory in the continuous configuration space fulfilling a symbolic goal and satisfying additional geometric constraints. We adopt an optimization-based formulation of Task-and-Motion Planning: a Logic-Geometric Program (LGP)\cite{Toussaint2015}\cite{Toussaint2018}. A feasible trajectory for a manipulation problem in this formulation consists of several phases connected by \textit{mode switches} \cite{Toussaint2018, Hauser2011}, for example, by actions (\textit{pick}, \textit{place}). %
A Logic-Geometric Program is a nonlinear mathematical program (NLP)
\begin{subequations}\label{eq:lgpall}
\begin{align} 
	\min_{K, s_{1:K} \atop x:[0,t_K]\to \XX } ~& \int^{t_K}_0 c(\bar x(t))~ dt \\
    \text{s.t.}~~ & ~~~~~~~~x(0) = x_0,~~~~\, h_\text{goal}(x(t_K))\leq 0, \\
    & \forall_k: \forall_t\in[t_{k\1}, t_k]: h_\text{mode}(\bar x(t), s_k) \leq 0, \\
    & ~~~~~~~~~~~~~~~\, h_\text{switch}(\bar x(t_k), s_{k\1}, s_k) \leq 0, \label{eq:lgp}\\
    & \forall_k: s_k \in \text{succ}(s_{k\1}), ~ s_K \models \mathfrak{g}
\end{align}
\end{subequations}

where $K$ is the number of phases, $\bar x(t)=(x(t), \dot x(t), \ddot x(t))$,  $s_{1:K}$ is the symbolic mode sequence, $x(t)$ is the continuous path, and $\mathfrak{g}$ is the logical goal described in first-order logic predicates.
The path is constrained to start at $x_0$ and end in a configuration $x(t_K)$ consistent with the goal constraints $h_\text{goal}$.
In each phase, the path needs to fulfill mode constraints $h_\text{mode}$ depending on the symbolic mode $s_k$, and each $\bar x(t_k)$ needs to fulfill mode-switch constraints $h_\text{switch}$ depending on the symbolic mode transition. 
While optimizing the smooth differentiable constraints within a phase is very efficient, solution search over different mode switches is a difficult combinatorial problem.
\subsection{Regrasp Maps for Sequential Manipulation TAMP}
In the presence of multiple regrasps, searching over many possible intermediate placements is often intractable. To address this, we propose a state space abstraction: a \emph{regrasp map} $M=\{M_{v}, M_{e}\}$, that captures which grasps are feasible in different regions of the configuration space and how those regions are connected via adjacency and regrasp sequences.

The map $M$ consists of \textit{grasp states} $M_v$ and connecting edges $M_e$. Every grasp state $v \in M_v$ is a tuple $(a, u)$ defined by the subspace of the collision free object poses $a \subset SE(3)$ and the grasp $u \in U_o$ from the set of available relative end-effector grasp poses $U_o \subset SE(3)$. 

$M$ is an undirected graph, and the edges indicate either \emph{transporting} the object with a given grasp or \emph{regrasping} in the same area with a different grasp. 
A path in $M$ is a sequence of grasp states $\{v_0 = (a^0, u^0), \ldots ,v_n = (a^n, u^n)\}$ with $v_i \in M_v$, where $v_i$ and $v_{i+1}$ are directly connected. 
We formulate our problem as finding such sequences in the regrasp map, the \emph{regrasp sequences}, and \emph{refining} them to a valid trajectory satisfying the symbolic goal of the TAMP problem. These sequences provide the solver with a guess for valid mode switches from the regrasps and with constraints on intermediate placements from the poses subspaces of the grasp states. %

\section{Method}
Our method consists of the following steps: grasp generation~(\ref{sec:grasps}), regrasp map construction and regrasp sequence search~(\ref{sec:map}), and a refinement phase~(\ref{sec:refine}), where we use and update the regrasp sequences to inform the TAMP solver.

\subsection{Grasps}
\label{sec:grasps}
Manipulation in tight spaces often requires continuous search over grasp parameters, as a purely discrete set of candidate grasps is typically impractical in such settings.

\begin{figure}[t]  
  \subfloat[Collision-free samples]{\label{fig:collisionfreegrasps}\includegraphics[width=0.20\linewidth, angle=90] {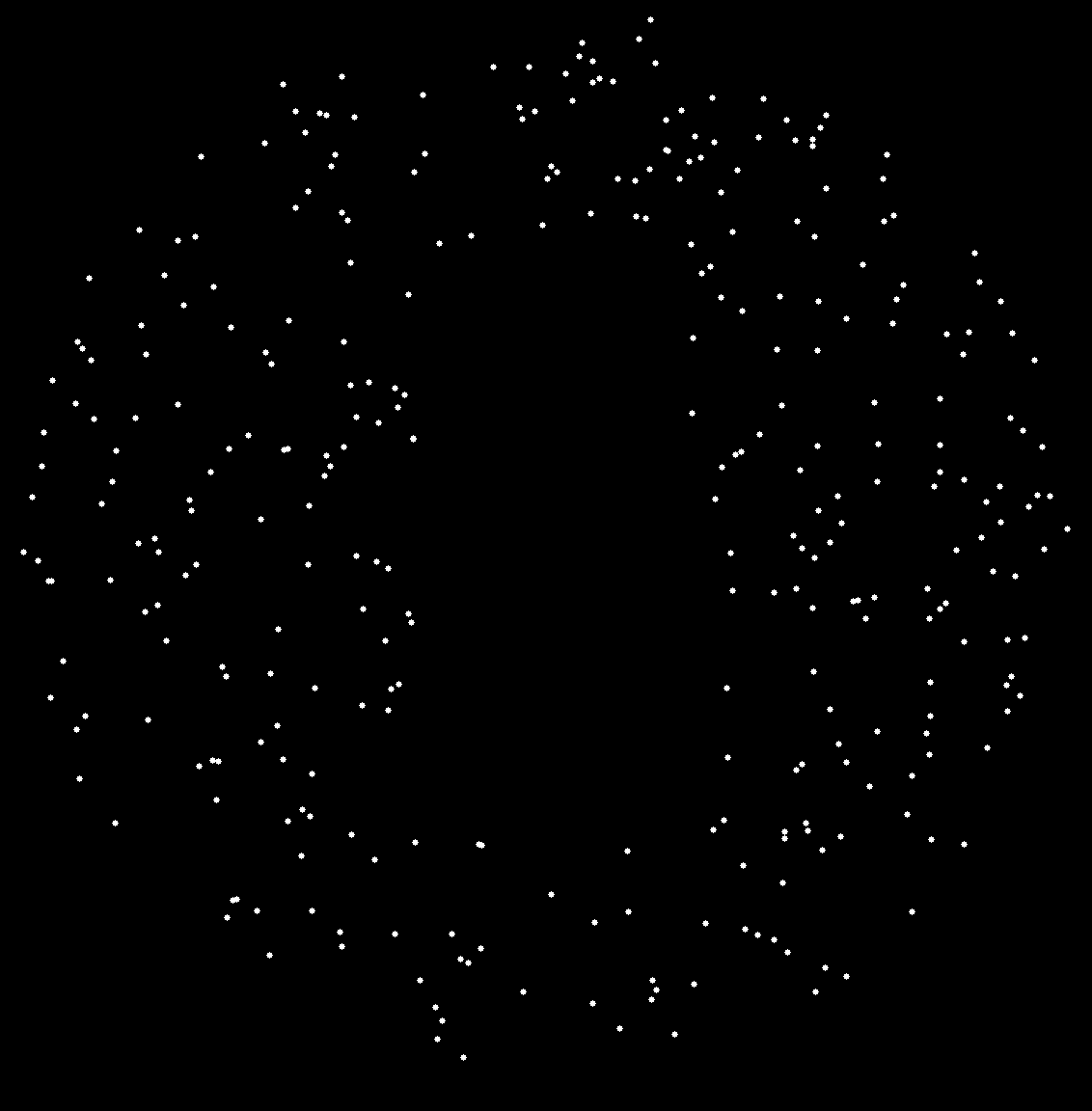}}\hfill
  \subfloat[Solution clusters]{\label{fig:graspclusters}\includegraphics[width=0.2\linewidth,angle=90]{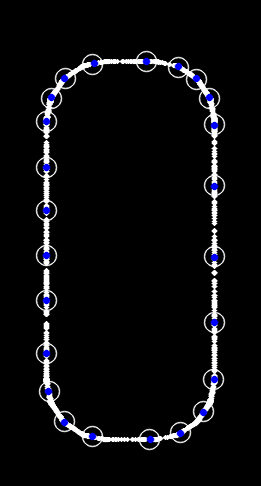}}\hfill
  \subfloat[Grasp anchors]{\label{fig:anchors}\includegraphics[width=0.2\linewidth, angle=90]{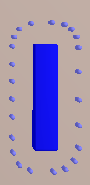}}\hfill
\caption{Generating K = 24 grasps by solving randomly initialized grasp problems and clustering the solutions. Blue dots in \ref{fig:graspclusters} are the positions closest to the center of the clusters, which we use as graph anchors.}
\label{fig:graspssearch}
\end{figure}

The optimization-based approach we employ allows us to overcome this issue by jointly optimizing the grasp pose as part of a nonlinear mathematical problem (NLP) together with any additional symbolical or geometrical constraints. To impose a geometric constraint on the grasps we define a set of \emph{grasp anchors} \{$u_1, \dots, u_k$\},
feasible reference poses of the end-effector relative to the object. We find them by solving multiple grasp problems for uniformly sampled positions around the object, then use clustering (Fig.~\ref{fig:anchors}) to select a representative set of anchors. For each anchor $u$, we sample approximate poses $u_{b}$ to check collision-free feasibility against the environment. A grasp’s \emph{feasibility score} $\phi$ reflects how many samples remain collision-free:
\begin{align}
   u \in U, \tilde{u} = [p_1 \ldots p_B], ~ \forall p : \| u - p \|_2 \leq \epsilon,  \\
   \phi(x_o, u) = \frac{\sum_{b \in [1..B]} \psi(x_o, \tilde{u}_b)}{B}
    \label{eq:graspsfeas}
 \end{align}
 A grasp is feasible if its \emph{feasibility score} $\phi$ is above a threshold $\alpha$. We compute $\psi$ using a combination of collision checks over hierarchical collision volumes and estimates for collision-free object placements using a Euclidean Signed Distance Field (ESDF) for the scene without the movable object and the robot.

Figure \ref{fig:graspssearch} visualizes the grasp generation using KOMO and k-means clustering.

\subsection{Constructing the Regrasp Map}
\label{sec:map}
\begin{figure}[t]
  \subfloat[Maze: white free space, brown walls]{\label{fig:grid}\includegraphics[width=0.323\linewidth]{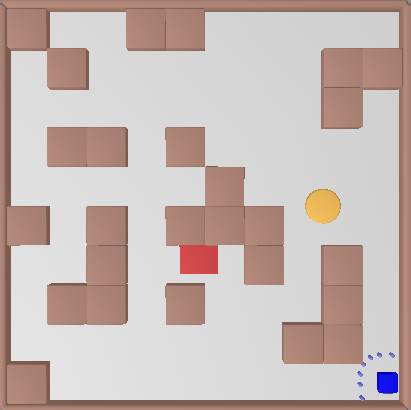}}\hfill
   \subfloat[Heat map of distances to closest obstacles (ESDF)]{\label{fig:pathsb}\includegraphics[width=0.323\linewidth]{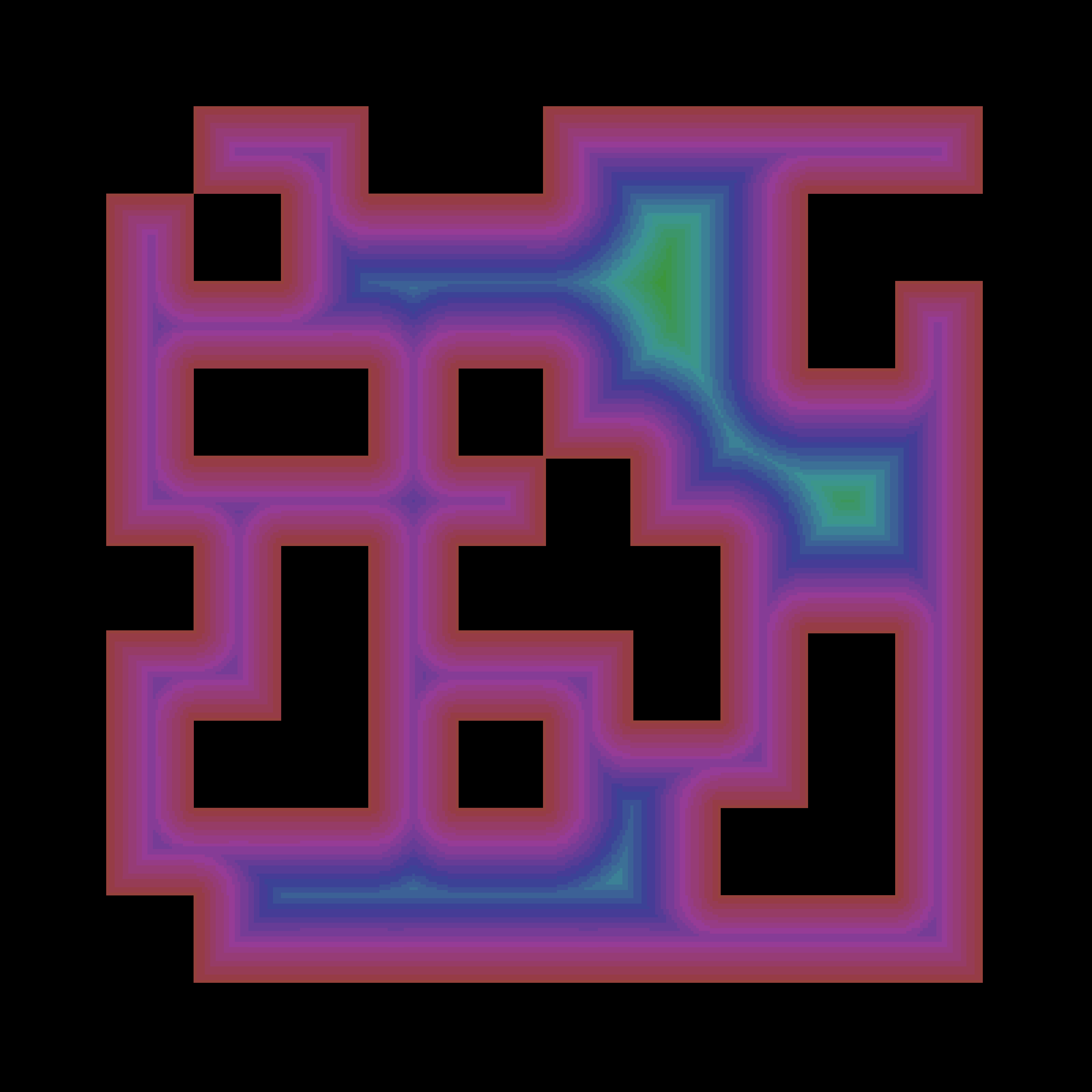}}\hfill
  \subfloat[Regrasp Map: black if has no valid grasps]{\label{fig:i10}\includegraphics[width=0.323\linewidth]{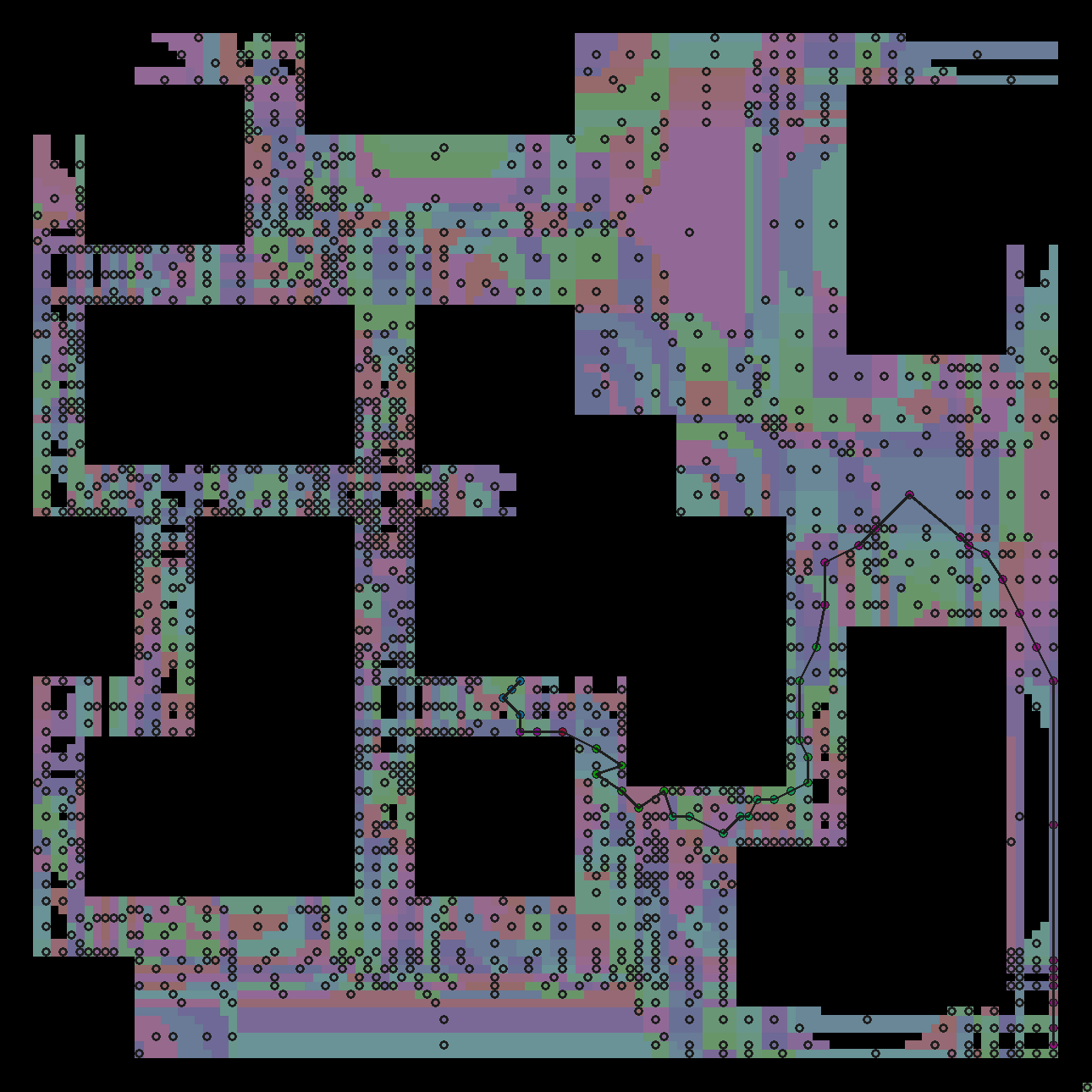}}\hfill
\caption{Segmenting the maze scene into areas with similar grasps (K=24 grasps).}
\label{fig:graspsareas2}
\end{figure}

To plan multiple regrasps, we partition the object’s configuration space $X_o$ based on \emph{which subset of grasps is feasible} there. 
Figure~\ref{fig:graspsareas2} shows a regrasp map segmentation for a maze scene: in the connected areas of the same color, the same grasps are possible. 
Algorithm~\ref{alg:plans} outlines the key steps to find regrasp sequences for a new scene, while algorithm~\ref{alg:map} describes in detail the creation of the regrasp map.

\paragraph{Segmentation}
First, we discretize $X_o$ into voxels (size $v$). For each voxel $q$, we compute the feasibility bit-string
\[
e(q)\in\{0,1\}^K,\quad 
[e(q)]_k=1 \;\;\text{if}\;\; \phi(q,u_k)\ge \alpha,
\]
which encodes which grasps $u_k$ are valid in $q$. We then group adjacent voxels with identical bit-strings into connected \emph{areas}. Thus, each area $a$ has a single \emph{grasp signature} identical to its voxels.

\subsubsection{State connections}
For each area $a$, we create a node $(a,u)$ in $M$ for every grasp $u$ indicated by $a$’s bit-string. We connect nodes if they share an area or share a grasp and belong to adjacent areas. Edge weights reflect the minimum feasibility scores,
\begin{equation}
w = - \log\bigl(\min( \phi(a, u), \phi(a', u'))- \varepsilon\bigr),~ \varepsilon \ll 1
  \label{eq:weights}
 \end{equation}

This results in an undirected graph representing feasible grasps and their transitions. Abstract paths in $M$ thus correspond to potential multi-step regrasp sequences.

\begin{figure}[t]   
  \subfloat[Areas partitioned by grasp signatures]{\label{fig:gridsamples}\includegraphics[width=0.39\linewidth]{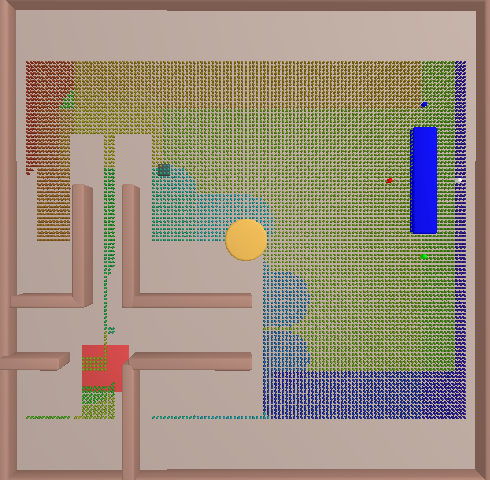}}\hfill
  \subfloat[\texttt{Tunnel} scene regrasp map for 4 grasps, with grasps in the same colors as markers around the blue object. Faint nodes are the rest of the graph.]{\label{fig:circles}\includegraphics[width=0.60\linewidth]{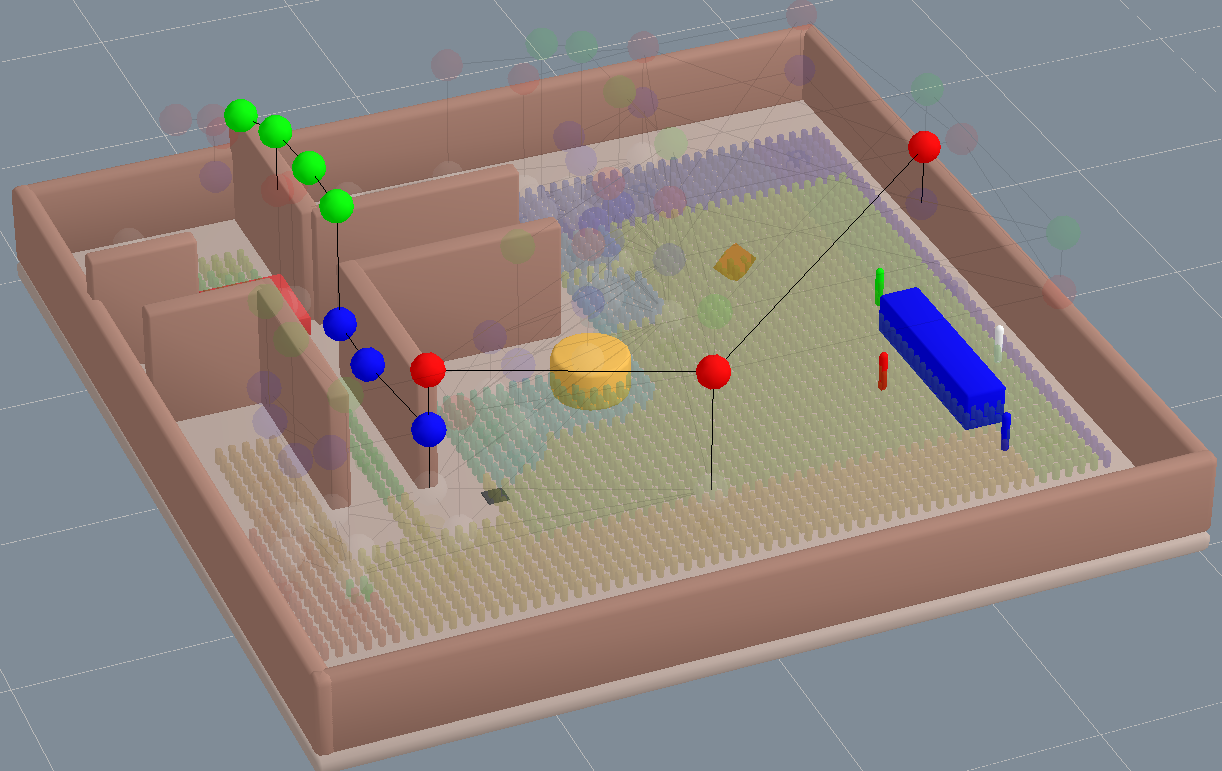}}\hfill
\caption{A regrasp sequence found for \texttt{Tunnel}: left grasp (red), then bottom grasp (blue), then top grasp (green). Vertical connections indicate regrasps inside the same area.}
\label{fig:stickmaps}
\end{figure}

Using the regrasp map, we can create abstract grasp plans to transport an object between any two positions in the scene.
Figure~\ref{fig:stickmaps} shows an example regrasp plan, where the agent (yellow) moves the blue object inside the enclosure using two regrasps.

\begin{algorithm}[h] %
  \caption{ Find Regrasp Plans}
\label{alg:plans}
\begin{algorithmic}[1] %
	\State \textbf{Input:} scene $x$, object $o$, voxel size $v_0$, logical goal $\mathfrak{g}$, number of grasps $K$, minimal voxel size $\gamma$
    \State $v \leftarrow v_0$, $P \leftarrow \emptyset$
     \State {$U_o \leftarrow$ \texttt{generate\_grasps($o$, $K$)}}
	\While {$P = \emptyset~ \wedge v > \gamma $} \Comment{While no path found}
    \State $D \leftarrow$ \texttt{grid($x$, $v$)}
	\State {$(M_v, M_e) \leftarrow$ \texttt{regrasp\_map(}$x$, $o$, $D$, $U_o$)} \Comment{Alg.~\ref{alg:map}}
	\State $P \leftarrow$ \texttt{find\_paths($x$, $(M_v$, $M_e$), $\mathfrak{g}$)}
    \State {$v \leftarrow v/2$} \Comment{Increase the grid resolution}	
	\EndWhile
	\Return $P$, $M$
  \end{algorithmic}
\end{algorithm}

\begin{algorithm}[h] %
  \caption{ Building a Regrasp Map}
\label{alg:map}
\begin{algorithmic}[1] %
	\State \textbf{Input:} scene $x$, object $o$, grid $D$, feasibility threshold $\alpha$, grasps $U_o$
	\For{$q \in D$}	\Comment{ For all query points in the grid}
	\If {\texttt{object\_free}($x$, $o$)}
	\For{$u \in U_o$} \Comment{To get grasp feasibility} 
	\State $\phi(q, u) \leftarrow$ \texttt{grasp\_free}$(q, u)$ \Comment{Eq.~\ref{eq:graspsfeas}}
	\EndFor    
    \EndIf
    \State {$G_q \leftarrow $ \texttt{grasp\_sgnt($\phi(q, [u_1, \dots , u_K]$))}} 
	\EndFor
    \State {$A \leftarrow $ \texttt{segment($G$)}} \Comment{Segment using signatures $G$}

	\For {$u \in  U_o , a \in A$} \Comment{For areas and grasps}
	  \State {$M_v \leftarrow M_v \cup (a, u)$ \Comment{New state for grasp $u$}}   
	\EndFor
    \For {$(a, u) \in M_v,~ (a', u') \in M_v$} 
    \If {$\phi$($a$, $u$) $>$ $\alpha~\wedge~\phi$($a'$, $u'$) $>$ $\alpha$ }
    \If {$a = a' \vee u = u'$} \Comment{Add an edge}
    \State {$w \leftarrow $ \texttt{weight($a, a', u, u'$)}} \Comment{Eq.~\ref{eq:weights}}
    \State {$M_e \leftarrow M_e \cup$ (($a$, $u$), ($a'$, $a'$), $w$)}
    \EndIf
    \EndIf
    
    \EndFor
  
\State \textbf{Return} $\texttt{$M_v$, $M_e$}$ \Comment{Return the regrasp map M }
  \end{algorithmic}
\end{algorithm}

\subsection{Refining the Regrasp Plans}
\label{sec:refine}

Using the regrasp map, we can integrate the dependencies between grasps and trajectories into our solver. Algorithm~\ref{alg:refine} outlines the refinement steps. We initialize a queue of configurations with our start scene $x_0$. Then, in \texttt{select\_node}, we determine which areas and states in the regrasp map $M$ are reachable from the configurations in $Q$ and select the ones with the lowest path cost in \texttt{select\_node}. For the selected configuration we first try to solve the initial LGP problem with Multi-Bound-Tree-Search (MBTS) LGP solver\cite{17-toussaint-ICRA}. If the solution was found, we reconstruct the solution from the last node to the start configuration. If the solver fails, we query the next graph state using the \texttt{next} method. If there is any path between the state $(a_x, u_x)$ and the goal, the method will return the state within one regrasp with the lowest path cost in $M$. Finally, the \texttt{use\_grasp} method attempts to reach the selected state: we formulate a 2-phase LGP problem with a fixed sequence of \texttt{pick} and \texttt{place} mode switches. For the grasp, in addition to the \textit{touch constraint}, we specify a \textit{pose constraint} using the grasp anchor position. For the placement, we use a location inside the selected area as a sub-goal pose constraint. We optimize jointly the grasp, trajectory, and the placement. 
If the solution is feasible, we add the final configuration of the solution trajectory as a new node.
The outcome of the search is passed to the \texttt{update} routine, where the edge weights are adjusted by increasing or decreasing the $\phi$ values of the attempted connection. Both \texttt{select\_node} and \texttt{next} query the distances of the shortest goal-directed paths in $M$, which are computed with Dijkstra shortest paths in \texttt{find\_paths} routine at the start (Algorithm~\ref{alg:plans}) and at each weights update. The search continues until either the $Q$ queue is empty or we find a valid solution. 

\begin{algorithm}[h] %
  \caption{Refine the Regrasp Plan}
\label{alg:refine}
\begin{algorithmic}[1] %
	\State \textbf{Input:} scene $x_0$, object $o$, regrasp map $M$, goal $\mathfrak{g}$%
    \State $Q \leftarrow x_0$ \Comment{Queue of nodes}
    \While {$\mathfrak{g}$ not satisfied (Eq.~\ref{eq:lgpall}) $\wedge$ $Q \neq \emptyset$}
      \State $x \leftarrow$ \texttt{select\_node($Q$, $M$)} \Comment{Next node}
    	\State $X \leftarrow \texttt{solve\_lgp}(x, \mathfrak{g})$ \Comment{Is logical goal reached?}
        \If {$X = \emptyset$}
        \State {$(a_x, u_x) \leftarrow $ \texttt{get\_state}($x$, $M$)}
    \State $a, u \leftarrow$ \texttt{next}($M$, ($a_x$, $u_x$)) \Comment{Next area \& grasp}
    \State $r$, \text{succ} $\leftarrow$ \texttt{use\_grasp(}$x$, $o$, ($a$, $u$))
  \State $Q \leftarrow$ $Q$ + $r_{end}$ \Comment{Add new states to the queue}
   \State \texttt{update(}$M$, $u$, $a$, \texttt{succ)}  \Comment{Update weights}
  \EndIf
    \EndWhile
\State \textbf{Return} $X$ \Comment{Return a valid trajectory}
  \end{algorithmic}
\end{algorithm}

\begin{table}[t] 
 \caption{Success rate (\%). For \texttt{r1-r4} over the subset, for single problems over 10 runs. \texttt{RMP}: regrasp map, Variations: \texttt{RMPfree}: without predefined grasps for the next step, \texttt{RMPfix}: without weight update and replan. Baselines: \texttt{RND}: forward search with random placements, \texttt{RNDh}: \texttt{RND} with visibilitiy heuristics\,}
 \label{tab:feas} 
  \begin{center} \begin{tabular}
{|l|c|c|c|c|c|}\hline 
\textbf{Test set (size)}&\textbf{RMP}&\textbf{RMPfix}&\textbf{RMPfree}&\textbf{RND}&\textbf{RNDh}\\ 

\hline 
Cage-Small (1)&\textbf{100}&\textbf{100}&\textbf{100}&\textbf{100}&\textbf{100}\\ 
\hline 
Cage (1)&\textbf{100}&\textbf{100}&\textbf{100}&-&-\\ 
\hline 
Tunnel (1)&\textbf{100}&\textbf{100}&\textbf{100}&-&-\\ 
\hline
r1 (36)&97.2&97.2&\textbf{100}&91.6&97.2\\ 
\hline 
r2 (37)&\textbf{86.4}&\textbf{86.4}&83.7&72.9&\textbf{86.4}\\ 
\hline 
r3 (18)&\textbf{89}&72.2&66.6&50&50\\ 
\hline 
r4 (9)&\textbf{77.7}&22.2&55.5&11.1&22.2\\ 
\hline 
  \end{tabular} 
 \end{center}
 \end{table}

\section{Experiments}
We evaluate our method using a 2-DoF robot agent moving in a continuous environment with static obstacles and a single movable object. The experiment scenarios involve manipulation in tight or cluttered spaces and require a sequence of alternating regrasps and moves in narrow spaces.

\begin{figure}[t]
\subfloat[Cage-Small]{\label{fig:i10}\includegraphics[width=0.325\linewidth]{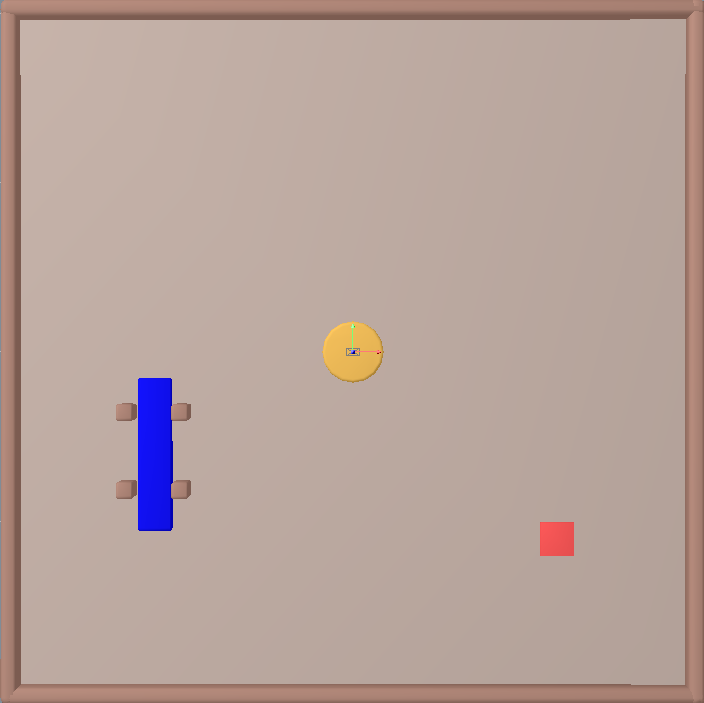}}\hfill
  \subfloat[Cage]{\label{fig:i10}\includegraphics[width=0.325\linewidth]{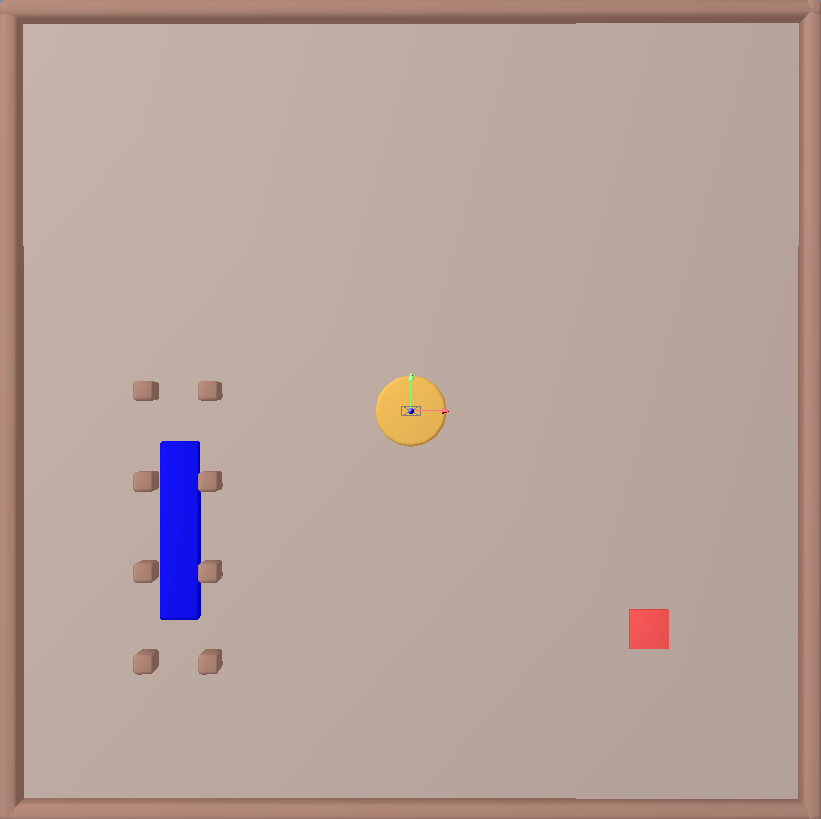}}\hfill
  \subfloat[Tunnel]{\label{fig:p1}\includegraphics[width=0.325\linewidth]{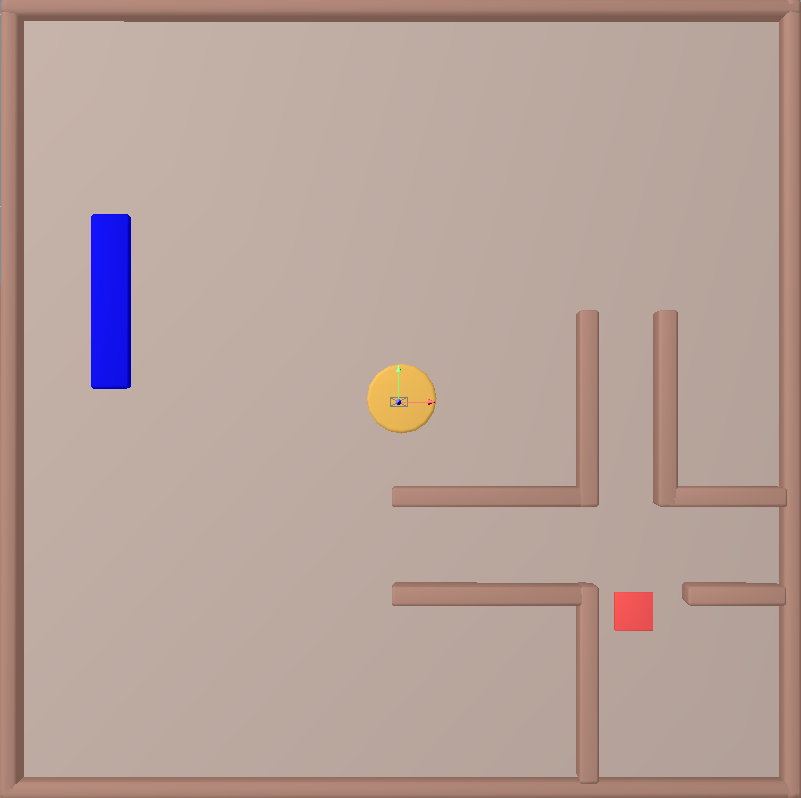}}\hfill
  \subfloat[Maze1]{\label{fig:r2}\includegraphics[width=0.325\linewidth]{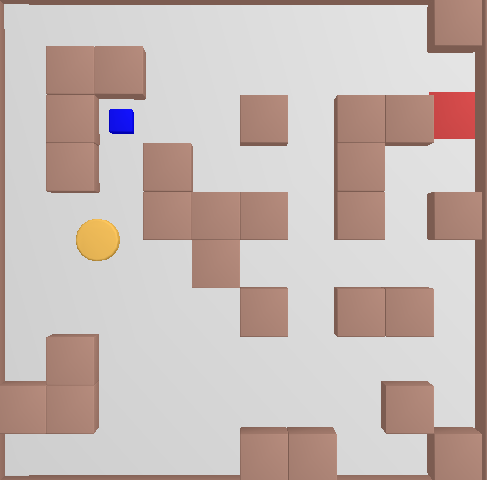}}\hfill
  \subfloat[Maze2]{\label{fig:r3}\includegraphics[width=0.325\linewidth]{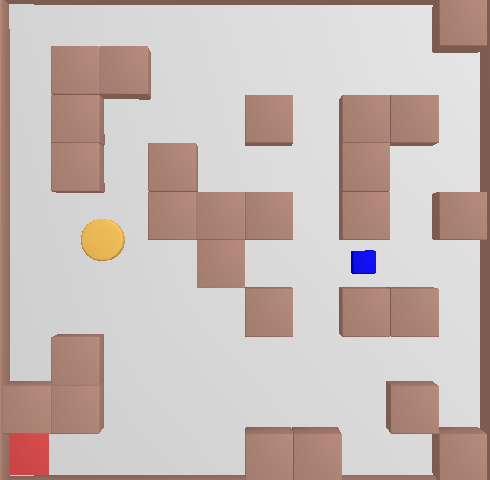}}\hfill
  \subfloat[Maze3]{\label{fig:r3}\includegraphics[width=0.325\linewidth]{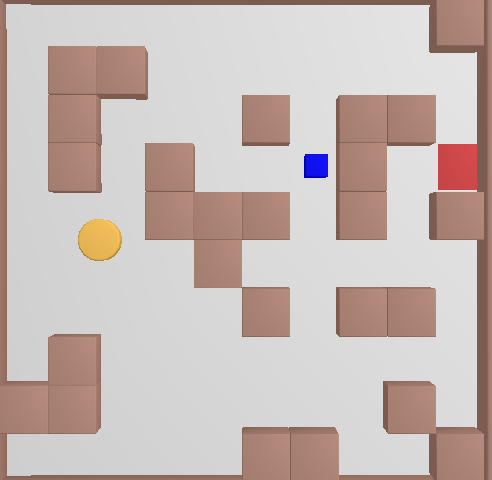}}\hfill
\caption{The agent (yellow circle) needs to bring the object (blue) to the goal (red square).}
\label{fig:problems3}
\end{figure}

Figure \ref{fig:problems3} shows the used setups (\texttt{Cage}, \texttt{Tunnel}) and examples of the \texttt{Maze} set.
To evaluate our method on multi-step regrasp tasks, we created a larger test set based on the \texttt{Maze} scene with randomized positions and separated scenes requiring regrasps into five subsets, depending on the number of regrasps in a shortest transport path. We refer to the subsets of this test set as \texttt{r1-r4}, according to the number of regrasps. 

The \texttt{Tunnel} requires a single regrasp, but at a precise location. The \texttt{Cage} scene is inspired by \cite{Simeon2004} and requires a sequence of alternating regrasps and object moves, while the \texttt{Cage-small} is an easier version, where no regrasp is needed, but the object movement and choice of grasps are limited.

For the problems we consider, it is necessary to plan with intermediate placements, while the number and locations of those placements are unknown.
We compare our \texttt{RMP}, regrasp map based planner, against baselines that performs forward search over intermediate placements: \texttt{RND} and a visibility-heuristic based \texttt{RNDh}\cite{24-levit-ICRA}.
\texttt{RND} is a more general version sampling subgoals not biased towards the goal and \texttt{RNDh} is filtering subgoals by their reachability and line-of-sight visibility towards the robot or the goal.

Furthermore, we evaluate variations of our \texttt{RMP} by removing components of our method:
\begin{itemize}
\item \texttt{RMPfix}: computes a fixed regrasp map at the start and selects the next action based on the adjacency of the states and their initially calculated distance to the goal.
This variation does not update the edge weights and has no replanning step.
\item \texttt{RMPfree}: same as \texttt{RMP}, but does not constrain which specific grasp to use, only the placement area.
\end{itemize}
We report the \textit{average solution time} and \textit{success rate} for individual scenes, including three scenes of the \texttt{Maze} setup, and \textit{success rates} over subsets of \texttt{Maze} problems (\texttt{r1-r4}). We performed the experiments on a single i7-11700@2.5\,GHz CPU. To speed up the collision checks during the map construction we pre-compute ESDF maps using \texttt{nvblox}\cite{Millane24} and a single NVIDIA RTX3060 GPU.

\section{Results}

 {\begin{table*}[t] 
 \caption{Average run time over 10 successful runs (s). Methods as in Tab.~\ref{tab:feas}. For \texttt{RMP} methods: the time \textbf{includes} the regrasp map construction, in brackets the refinement time. '-' if the problem was not solved}
 \label{tab:time} 
 \begin{center} \begin{tabular}{|l|c|c|c|c|c|}\hline 
\textbf{Scene}&\textbf{RMP}&\textbf{RMPfix}&\textbf{RMPfree}&\textbf{RND}&\textbf{RNDh}\\ 
\hline
Cage-Small&26.59{\tiny $\pm$0.13}~~[12.31{\tiny $\pm$0.05}]&27.78{\tiny $\pm$0.22}~~[13.50{\tiny $\pm$0.12}]&25.62{\tiny $\pm$0.21}~~[11.34{\tiny $\pm$0.10}]&4.92{\tiny $\pm$4.19}&\textbf{2.92}{\tiny $\pm$2.24}\\ 
\hline 
Cage&106.27{\tiny $\pm$0.74} [38.22{\tiny $\pm$0.23}]&149.47{\tiny $\pm$2.06} [80.76{\tiny $\pm$1.25}]&\textbf{101.26}{\tiny $\pm$1.49}~[30.87{\tiny $\pm$0.35}]&-&-\\ 
\hline 
Tunnel&94.32{\tiny $\pm$0.86}~~[34.43{\tiny $\pm$0.50}]&138.01{\tiny $\pm$0.51} [78.34{\tiny $\pm$0.21}]&\textbf{83.78}{\tiny $\pm$0.29}~~[24.33{\tiny $\pm$0.07}]&-&-\\ 
\hline 
Maze1&\textbf{30.38}{\tiny $\pm$0.17}~[17.37{\tiny $\pm$0.12}]&93.01{\tiny $\pm$0.72}~[79.92{\tiny $\pm$0.67}]&37.84{\tiny $\pm$0.21}~[24.83{\tiny $\pm$0.14}]&229.82{\tiny $\pm$26.65}&211.25{\tiny $\pm$25.93}\\ 
\hline
Maze2&\textbf{49.16}{\tiny $\pm$0.30}~[36.13{\tiny $\pm$0.20}]&-&135.08{\tiny $\pm$0.66}~[122.09{\tiny $\pm$0.64}]&-&974.43{\tiny $\pm$484.51}\\ 
\hline
Maze3&36.78{\tiny $\pm$0.11}~[23.76{\tiny $\pm$0.09}]&97.80{\tiny $\pm$0.37}~[84.75{\tiny $\pm$0.36}]&\textbf{30.62}{\tiny $\pm$0.16}~[17.58{\tiny $\pm$0.09}]&-&594.35{\tiny $\pm$407.58}\\ 
\hline
 \end{tabular} 
 \end{center}
 \end{table*}}

\begin{figure}
\centering
\includegraphics[width=1.0\linewidth]{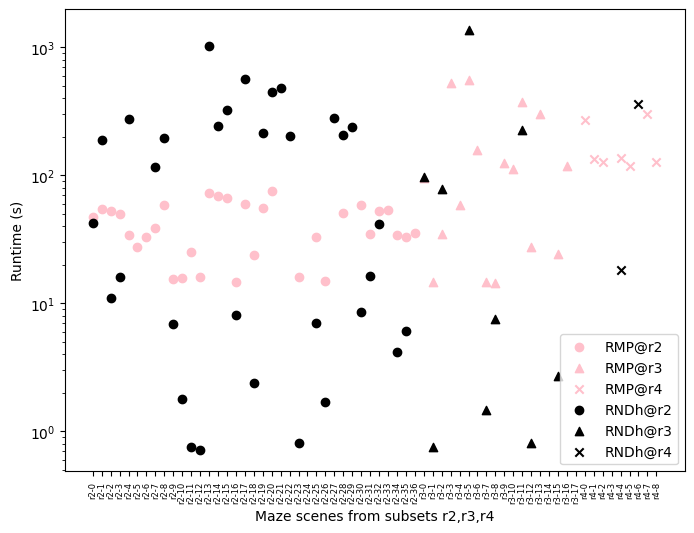}
	\caption{Solution times for successful runs on subsets of maze scenes with 2-4 regrasps: r2, r3, r4 (log scale)} 
    \label{fig:scatter}
\end{figure}

In Table \ref{tab:feas} we report the success rate for all scenes and in Table~\ref{tab:time} the average time to solution for the individual scenes. 
All methods perform well on \texttt{Cage-Small} and scenes with a single regrasp (set \texttt{r1}).
For the set \texttt{r4} the methods \texttt{RMP} and \texttt{RMPfree} solve the most problems.
We attribute this to the replanning step, which allows to pick a different placement if the refinement of the initially computed connections fails.
Especially for multi-step problems it is more likely that the plans calculated initially will not contain the states reached or that one of the sub-problems could not be solved.
In comparison, a failed refinement in \texttt{RMPfix} leads to the planner retrying the same connections later, increasing the runtime.  
The \texttt{RMPfree} variation shows higher success rate when the knowledge about the grasp feasibility is incorporated into the regrasp map:
The \texttt{RMP} can solve almost all problems with a higher success rate than \texttt{RMPfree}, while having a similar runtime.
In \texttt{RMPfree} we only specify which placements to use, but not which grasp (\textit{i.e.}, no \textit{pose constraint} is added to the \texttt{pick} step). For problems with several regrasps, \texttt{RMPfree} has a lower success rate due to a narrow range of grasps which would lead to a solution.
However, for problems with narrow passages as \texttt{Tunnel}, where the location itself imposes constraints on the possible grasps, the \texttt{RMPfree} performs as well as the grasp-informed \texttt{RMP}.
For the \texttt{r1-r4} sets we show the runtimes of individual problems in Fig.~\ref{fig:scatter} instead of the averages, as the distribution of solution times for the scenes in sets is rather non-Gaussian.
In terms of runtime, \texttt{RMP} has a higher computation overhead, due to computation of the regrasp maps and grasp generation. 
However, for scenes with more regrasps, the solution time is comparable or better than \texttt{RNDh}, while having a consistently better success rate.
Fig.~\ref{fig:scatter} shows the distribution of runtimes for the problems requiring between two and four regrasps.
While the solution time of \texttt{RMP} is longer for easier problems, particularly because of the map computation overhead, with longer regrasp sequences the total time is still shorter and has smaller spread. %

\section{Conclusion}
Our method leverages \emph{regrasp maps} to guide an optimization-based TAMP planner in solving complex manipulation problems with multiple regrasps. We demonstrate that incorporating regrasp information consistently enables the solver to handle difficult tasks,  outperforming alternatives that do not use such guidance. The advantage is especially noticeable in scenes with longer sequences of steps and requiring precise grasps and placements, where interleaved map updates help to achieve a feasible solution by adapting the plans after failed sub-problems. Although the overall runtime can be high, a large portion is spent on constructing the regrasp map, which can be precomputed or parallelized. Overall, these results confirm the effectiveness of integrating information about grasps and regrasps into long-horizon manipulation planning.

\bibliographystyle{IEEEtran}
\balance
\bibliography{IEEEabrv,bib/general.bib}

\end{document}